\pdfoutput=1

\documentclass[11pt]{article}

\usepackage[preprint]{acl}

\usepackage{times}
\usepackage{latexsym}

\usepackage[T1]{fontenc}

\usepackage[utf8]{inputenc}

\usepackage{microtype}

\usepackage{inconsolata}

\usepackage{graphicx}
\usepackage{CJKutf8}
\usepackage{amsmath}
\usepackage{booktabs}
\usepackage{multirow}
\usepackage{makecell}
\usepackage{colortbl}
\usepackage{tcolorbox}
\usepackage[tikz]{bclogo}

\newtcolorbox{myquote}{
    colback=gray!10, 
    colframe=gray, 
    fonttitle=\bfseries, 
    coltitle=black, 
    sharp corners, 
    boxrule=1pt, 
    width=\linewidth, 
    left=10pt, 
    right=10pt, 
    top=10pt, 
    bottom=10pt, 
}

\definecolor{my_blue}{RGB}{0,120,255}
\definecolor{my_purple}{RGB}{161, 27, 155}
\definecolor{my_green}{RGB}{0, 176, 80}
\definecolor{msftBlue}{RGB}{0,164,239}
\definecolor{msftGreen}{RGB}{127,186,0}
\definecolor{msftYello}{RGB}{255,185,0}
\definecolor{msftBlack}{RGB}{0,0,0}
\newcommand{\finding}[1]{
\begin{bclogo}[couleur= msftBlack!05, epBord=1, arrondi=0.1, logo=\bclampe,marge=2, ombre=true, blur, couleurBord=msftBlack!10, tailleOndu=3, barre = none, sousTitre ={\em #1}]{} 
\vspace{0.5em}
\end{bclogo}
}

\title{Memorizing is Not Enough: Deep Knowledge Injection Through Reasoning}

\author{Ruoxi Xu$^{1,2}$, Yunjie Ji$^{3}$, Boxi Cao$^{1}$, Yaojie Lu$^{1}$, Hongyu Lin$^{1}$, Xianpei Han$^{1}$,\\ \textbf{Ben He$^{2}$, Yingfei Sun$^{2}$, Xiangang Li$^{3}$, Le Sun$^{1}$}\\
$^{1}$Chinese Information Processing Laboratory, Institute of Software, Chinese Academy of Sciences\\$^{2}$University of Chinese Academy of Sciences, $^{3}$ \includegraphics[width=0.4cm]{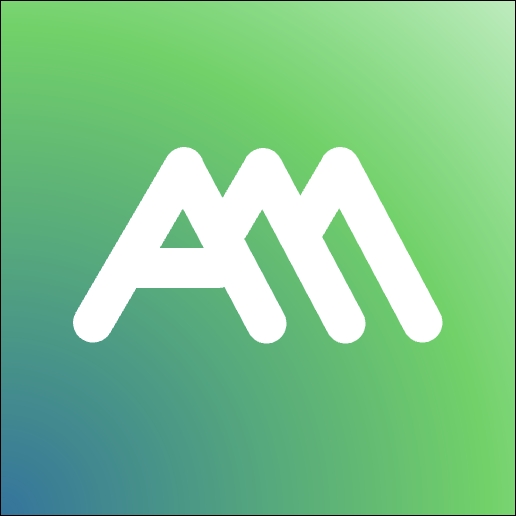} a-m-team\\
}

\begin{document}
\maketitle
\begin{abstract}
Although large language models (LLMs) excel in knowledge recall and reasoning, their static nature leads to outdated information as the real world evolves or when adapting to domain-specific knowledge, highlighting the need for effective knowledge injection.
However, current research on knowledge injection remains superficial, mainly focusing on knowledge memorization and retrieval.
This paper proposes a four-tier knowledge injection framework that systematically defines the levels of knowledge injection: memorization, retrieval, reasoning, and association.
Based on this framework, we introduce DeepKnowledge, a synthetic experimental testbed designed for fine-grained evaluation of the depth of knowledge injection across three knowledge types (novel, incremental, and updated).
We then explore various knowledge injection scenarios and evaluate the depth of knowledge injection for each scenario on the benchmark.
Experimental results reveal key factors to reach each level of knowledge injection for LLMs and establish a mapping between the levels of knowledge injection and the corresponding suitable injection methods, aiming to provide a comprehensive approach for efficient knowledge injection across various levels.
\end{abstract}

\section{Introduction}

LLMs have the remarkable ability to capture vast amounts of factual knowledge from extensive pre-training data~\citep{alkhamissi2022review, cao2021knowledgeable, meyer2023llm}. However, their static nature leads to  knowledge becoming outdated as real-world information evolves or when adapting to new and private domain knowledge~\citep{wang2024knowledge}. To mitigate these issues, continual pre-training on updated or domain-specific documents has become a common strategy~\citep{zhang2023large, jang2022towards}, aiming to refresh LLMs' knowledge and tailor it to specific areas of expertise.

Knowledge injection progresses as a continuum, not a binary transition~\citep{hu2023survey}.
However, current research on remains superficial, mainly focusing on knowledge memorization and retrieval.
For instance,~\citet{carlini2021extracting, cao2024retentive} assess knowledge memorization through text completion tasks and~\citet{chang2024large, allenphysics} examine retrieval via rephrased questions.
Superficial knowledge struggles to support reasoning tasks, leading to under-performance in scenarios that require deep reasoning~\citep{allen2023physics}. Therefore, it is critical to conduct a systematic investigation about knowledge injection levels and establish a mapping between injection levels and suitable knowledge injection methods.

\begin{figure}[t]
    \centering
    \setlength{\belowcaptionskip}{-10pt}
    \includegraphics[width=\linewidth]{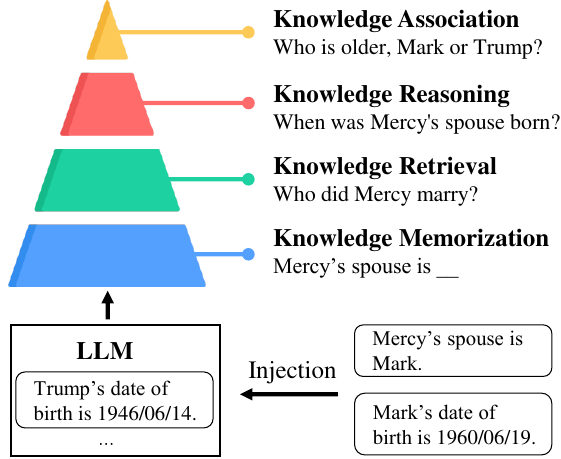}
    \caption{An illustration of the four-layer knowledge injection framework. This hierarchical framework provides a finer-grained approach to injecting knowledge into LLMs, ranging from basic recall to joint reasoning between new knowledge and pre-existing knowledge.}
    \label{fig:iceberg}
\end{figure}

To this end, this paper proposes a four-layer knowledge injection framework, systematically defining the four key levels of knowledge injection. As illustrated in Figure~\ref{fig:iceberg}, we divide the knowledge injection process into four levels: 1) \textbf{Knowledge Memorization}: The model's ability to recall and restate the injected knowledge in its original form. 2) \textbf{Knowledge Retrieval}: The model's ability to correctly extract knowledge under various expressions. 3) \textbf{Knowledge Reasoning}: The model's ability to apply the injected knowledge in reasoning tasks. 4) \textbf{Knowledge Association}: The model's ability to jointly apply the injected knowledge and pre-existing knowledge in reasoning tasks.

To investigate the interactions between these layers, we develop a synthetic experimental testbed called DeepKnowledge. DeepKnowledge offers a four-tier evaluation of knowledge injection effectiveness, based on four distinct knowledge types. Specifically, the evaluation aligns with the knowledge injection levels in Figure~\ref{fig:iceberg}: memorization (Level 1), retrieval (Level 2), 1-3 step reasoning (Level 3) and association (Level 4). This hierarchical evaluation allows for a more nuanced understanding of the challenges involved in integrating knowledge into LLMs, from basic recall to complex multi-step reasoning. Besides, DeepKnowledge incorporates four knowledge types: pre-existing, novel, incremental and updated knowledge. Knowledge is unique, and non-recursive to ensure valid multi-step reasoning.

Based on this setup, we systematically investigate the boundaries of knowledge injection under various knowledge injection scenarios. Our findings reveal key factors influencing the achievement of each level of knowledge injection in LLMs: 1) Repetitive learning enables rapid memorization of isolated knowledge; 2) Knowledge diversity is critical for transitioning from mere memorization to retrievable knowledge representation; 3) Explicit reasoning patterns link isolated knowledge for reasoning and enable generalization to new entities and deep reasoning; 4) LLMs excel at shallow knowledge association, but require explicit reasoning to forge deep connections. We infer that new knowledge must be interconnected through reasoning mechanisms to facilitate generalized knowledge reasoning.

Furthermore, we conducted ablation experiments to identify the key factors affecting the efficiency of knowledge injection. We analyzed the impact of knowledge types, data formulation and diversity on the effectiveness of knowledge injection and provided a recipe to achieve efficient knowledge injection at various levels, which could be valuable for future research. 

To summarize, we make the following contributions:
\begin{itemize}
    \item We proposed a four-layer knowledge injection framework, including knowledge memorization, retrieval, reasoning and association.
    \item We developed DeepKnowledge, a comprehensive evaluation testbed designed to assess knowledge application across different injection levels.
    \item We established a systematic mapping between knowledge injection levels and suitable injection methods.
\end{itemize}

\section{DeepKnowledge}

In this section, we present a comprehensive experimental testbed, DeepKnowledge. Aligned with the knowledge injection framework in Figure~\ref{fig:iceberg}, DeepKnowledge enables systematic evaluation across four knowledge injection levels. The following provides a detailed description of the construction process of DeepKnowledge.

\subsection{Knowledge Acquisition}

\paragraph{Pre-existing Knowledge Filter}

To ensure the validity of our benchmark, we apply three filtering criteria to pre-existing factual knowledge: 1) Uniqueness: For each subject-relation pair, the object must be unique. 2) Non-recursiveness: Facts should not be recursive, i.e., the subject and object cannot be identical. 3) Multi-step Reasoning: The knowledge must be suitable for constructing multi-step reasoning tasks.

Specifically, we first get factual knowledge from WikiFactDiff~\citep{khodja2024wikifactdiff} and MQuAKE~\citep{zhong2023mquake}. We then filter out facts containing special characters or empty values in the subject or object, as well as those with non-unique answers or recursive structures. This ensures that only high-quality, valid facts remain in the dataset. Next, we analyze the distribution of fact chains and manually select 16 relationship groups that are critical for reasoning tasks, as illustrated in Appendix Figure~\ref{fig:relation}. Finally, we enable the model to recall facts in a 3-shot setting, retaining only the facts that the model provides correct answers as pre-existing knowledge. In total, we curate a set of 26,477 valid and reliable facts.

\paragraph{Synthetic Knowledge Generation}

To ensure that the injected knowledge contains novel information that was not seen during the pretraining phase, we construct a synthetic knowledge dataset. To eliminate confounding factors, the synthetic knowledge is designed to match the distribution of factual knowledge. Specifically, based on the relationship types selected from the factual knowledge, we first generate fictional entity names using LLMs. For example, when generating names for entities belonging to the category of "location," we combine a random word (e.g., "Frank") with a geographic term (e.g., "town") to create an entity name such as "FrankTown." These fictional entities are then assigned the same relationships as those in the factual knowledge to form synthetic facts. As a result, we generate a set of 109860 synthetic facts.

\subsection{Test Cases Generation}

\paragraph{Memorization Test Cases Generation}

We define the shallowest level of knowledge injection as the model's ability to retain and recognize the original content of the knowledge during training. Specifically, we focus on sentences from the training corpus that involve specific knowledge. For these sentences, the object is removed, transforming the sentence into a cloze-style question. The task for the model is to predict the correct object, which serves as the answer to the cloze question. This method effectively measures the model's ability to recall and reconstruct knowledge based on its training data, providing a baseline for evaluating its memory and knowledge retention capabilities.

\paragraph{Retrieval Test Cases Generation}

We define knowledge extraction as the model's ability to extract knowledge from various semantically equivalent formulations. Specifically, we take the cloze questions from memorization-level test cases and rephrase them into ten semantically equivalent questions using LLMs. These rephrased questions are then used as the input, with the original answer retained as the correct response. This approach assesses the model's capacity to recognize and extract knowledge across different expressions while ensuring that the underlying semantic content remains consistent.

\begin{figure}[t]
    \centering
    \setlength{\belowcaptionskip}{-10pt}
    \includegraphics[width=\linewidth]{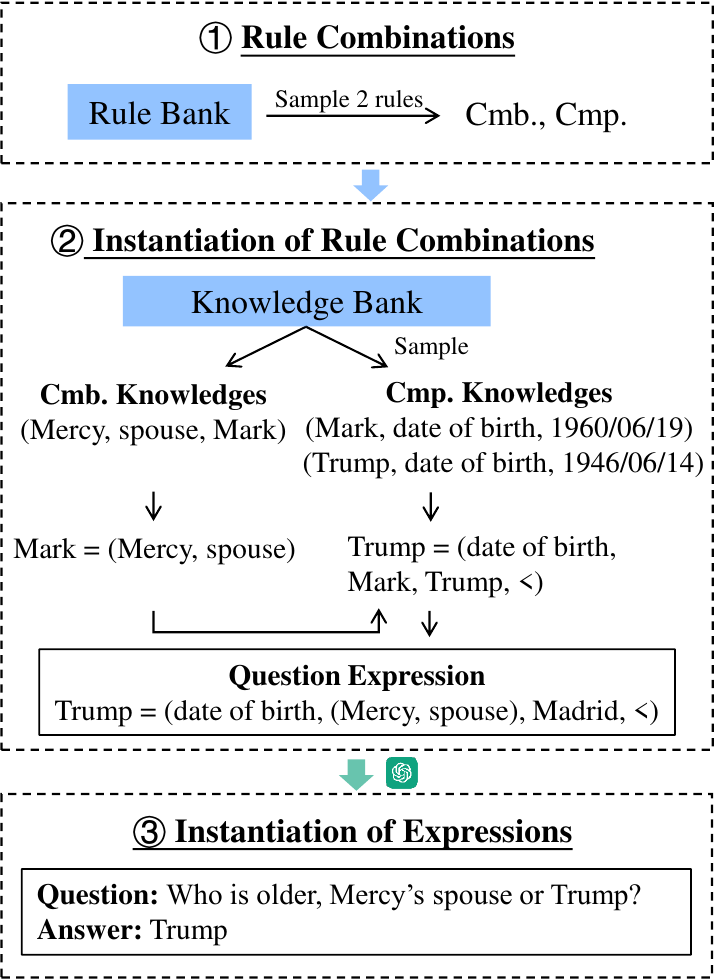}
    \caption{Construction pipeline for reason test cases, which involves three steps: 1) multi-step rule combinations, 2) instantiation of rule combinations to knowledge expressions following the rule definition, 3) instantiation of knowledge expressions to questions using LLMs.}
    \label{fig:pipeline}
\end{figure}

\paragraph{Reasoning Test Cases Generation}

We define reasoning as the induction and application of inference rules. Specifically, we first define two fundamental reasoning rules: combination, which involves multi-hop knowledge aggregation, and comparison, which focuses on comparing the magnitude of knowledge. Each reasoning rule is considered a single step of reasoning. To construct an n-step reasoning problem, we sample n reasoning rules as the foundation of the problem, and randomly sample knowledge to populate each reasoning step to form an expression. Finally, the expression is translated into a fluent natural language question using GPT-4. The construction pipeline for reason test cases is illustrated in Figure~\ref{fig:pipeline}. The detailed rule definitions and prompts can be found in the Appendix~\ref{sec:benchmark}.

\paragraph{Association Test Cases Generation}

We define association as the reasoning process that connects newly injected knowledge with pre-existing knowledge in LLMs. Specifically, we adopt an approach similar to question generation for reasoning tasks, with the key distinction that the constructed questions must incorporate both new and existing knowledge.

\section{Experiment Settings}

\subsection{Knowledge Types}

The correlation between newly injected knowledge and the pre-existing knowledge in LLMs may influence the effectiveness of the injection. Therefore, we systematically explore three knowledge paradigms that encompass possible types of new knowledge~\citep{khodja2024wikifactdiff}: 1) Novel Knowledge, which introduces entirely new information about emerging entities (e.g., a newly proposed scientific theory); 2) Incremental Knowledge, which expanding existing entities with supplemental facts (e.g., new publications by established authors); and 3) Updated Knowledge, where outdated facts are replaced with current information (e.g., a sports team appointing a new coach).

\subsection{Injection Scenarios}

We explored five distinct knowledge injection scenarios: 1) Duplicate: Repeating the same knowledge multiple times without modification. 2) Vanilla Paraphrase: Rephrasing the injected knowledge using a LLM, altering its expression. 3) Style-enhanced Paraphrase: Rephrasing the injected knowledge with style variations, where the rephrasing style is randomly selected from a pre-defined style bank (details provided in Appendix~\ref{sec:setting}). 4) Single-step Implicit Reasoning: Combining the rephrased knowledge with a single-step reasoning question and its corresponding answer. 5) Single-step Explicit Reasoning: Combining the rephrased knowledge with a single-step reasoning question, a detailed reasoning process, and the corresponding answer.
For all these knowledge injection scenarios, we ensured that each knowledge was injected 20 times, regardless of its form, to eliminate any potential influence of training data size.

\subsection{Test Settings}

We evaluate the model's performance under three distinct in-context learning settings: 1) 0-shot, where LLMs generate answers without any prior context or examples. 2) 3-shot, where three relevant examples are selected based on various criteria: for memorization, examples share the same relation type as the test question; for retrieval, examples contain answers related to the same entity type as the test question; and for reasoning and association, examples exhibit similar reasoning structures to the test question. 3) 3-shot CoT: which mirrors the 3-shot setting, but includes examples that explicitly demonstrate reasoning processes.

\subsection{Evaluation Metrics}

To evaluate the model's performance in answering free-text questions, we leverage tailored regular expression-based parsing techniques. Specifically, for memorization and retrieval questions, the model's response is deemed correct if the ground truth is contained within the first sentence of its generated text. For reasoning and association questions, we employ a prompt beginning with "Answer:" to guide the model in providing explicit options, extracting the initials of its generated answers. If these align fully with the ground truth, the response is considered accurate.

\subsection{Training Details}

We adopted continued pretraining (CPT) to inject new knowledge into LLMs, motivated by existing research that suggests supervised fine-tuning may induce hallucinations~\citep{gekhman2024does}. Our primary experiments were conducted on the LLaMA 3-8B model. To ensure stability and robust generalization, we trained LLMs using a balanced mixture of training data and general instructions at a 1:1 ratio, with a learning rate of 3e-5.

\begin{table*}[t]
\centering
\begin{tabular}{c|c|ccc|ccc}
\toprule
 &  & \multicolumn{3}{c}{Reason (2 steps)} & \multicolumn{3}{c}{Reason (3 steps)} \\
\multirow{-2}{*}{Injection Scenario} & \multirow{-2}{*}{Test} & \multicolumn{1}{c}{0S} & \multicolumn{1}{c}{3S} & \multicolumn{1}{c}{3S-CoT} & \multicolumn{1}{c}{0S} & \multicolumn{1}{c}{3S} & \multicolumn{1}{c}{3S-CoT} \\ \toprule
- & Old & \cellcolor[HTML]{FCFEFC}35.7 & \cellcolor[HTML]{F5FBF6}37.0 & \cellcolor[HTML]{63BE7B}64.0 & \cellcolor[HTML]{FDD5D6}26.3 & \cellcolor[HTML]{FEECEC}31.0 & \cellcolor[HTML]{92D2A3}55.3 \\ \midrule
 & Novel & \cellcolor[HTML]{F98E90}11.3 & \cellcolor[HTML]{FCCECE}24.7 & \cellcolor[HTML]{F8696B}3.3 & \cellcolor[HTML]{F98082}8.3 & \cellcolor[HTML]{FDD7D8}26.7 & \cellcolor[HTML]{F86A6C}3.7 \\
\multirow{-2}{*}{Duplicate} & Updated & \cellcolor[HTML]{F98385}9.0 & \cellcolor[HTML]{FBB0B1}18.3 & \cellcolor[HTML]{F86D6F}4.3 & \cellcolor[HTML]{F87B7D}7.3 & \cellcolor[HTML]{FBB6B7}19.7 & \cellcolor[HTML]{F8696B}3.3 \\ \midrule
 & Novel & \cellcolor[HTML]{FCC3C3}22.3 & \cellcolor[HTML]{FCD2D3}25.7 & \cellcolor[HTML]{FEEDED}31.3 & \cellcolor[HTML]{FDDFDF}28.3 & \cellcolor[HTML]{FDE4E4}29.3 & \cellcolor[HTML]{FCCECE}24.7 \\
\multirow{-2}{*}{Style-enhanced   Paraphrase} & Updated & \cellcolor[HTML]{FCCACB}24.0 & \cellcolor[HTML]{FDD5D6}26.3 & \cellcolor[HTML]{FDE8E9}30.3 & \cellcolor[HTML]{FCD1D1}25.3 & \cellcolor[HTML]{FBB7B8}20.0 & \cellcolor[HTML]{FEECEC}31.0 \\ \midrule
 & Novel & \cellcolor[HTML]{FDE0E1}28.7 & \cellcolor[HTML]{EAF7ED}39.0 & \cellcolor[HTML]{FEFBFB}34.3 & \cellcolor[HTML]{DCF1E1}41.7 & \cellcolor[HTML]{FEFAFA}34.0 & \cellcolor[HTML]{FEEFEF}31.7 \\
\multirow{-2}{*}{Single-step Implicit Reason} & Updated & \cellcolor[HTML]{FCCECE}24.7 & \cellcolor[HTML]{FDD7D8}26.7 & \cellcolor[HTML]{FEFFFE}35.3 & \cellcolor[HTML]{EEF8F0}38.3 & \cellcolor[HTML]{E5F4E9}40.0 & \cellcolor[HTML]{FEF8F8}33.7 \\ \midrule
 & Novel & \cellcolor[HTML]{FBBEBF}21.3 & \cellcolor[HTML]{FBBEBF}21.3 & \cellcolor[HTML]{DFF2E4}41.0 & \cellcolor[HTML]{FBBBBC}20.7 & \cellcolor[HTML]{FCCACB}24.0 & \cellcolor[HTML]{B2DFBE}49.3 \\
\multirow{-2}{*}{Single-step Explicit Reason} & Updated & \cellcolor[HTML]{FEF8F8}33.7 & \cellcolor[HTML]{FDE8E9}30.3 & \cellcolor[HTML]{A4D9B2}52.0 & \cellcolor[HTML]{FCCECE}24.7 & \cellcolor[HTML]{FDE7E7}30.0 & \cellcolor[HTML]{87CD9A}57.3 \\
\bottomrule
\end{tabular}
\caption{The knowledge reasoning score of LLMs across various training and test settings. "0S", "3S" and "3S-CoT" correspond to the 0-shot, 3-shot, and 3-shot CoT test settings, respectively. Scores are presented on a green-white-red scale. From the table, we can observe that newly injected knowledge is often isolated and needs to be connected through explicit reasoning, enabling generalization to new entities and deep reasoning.}
\label{tab:complex_reason}
\end{table*}

\section{What's the key to reach each level of knowledge injection for LLMs?}

In this section, we explore various knowledge injection scenarios and evaluate their knowledge injection levels on the DeepKnowledge testbed. Our experiments reveal that different knowledge injection strategies are required to attain various levels of knowledge injection in LLMs. In the following, we will illustrate the experiment findings to reach the above conclusion.

\begin{table*}[t]
\centering
\setlength{\tabcolsep}{4pt}
\begin{tabular}{ccllllll}
\toprule
 &  & \multicolumn{3}{c}{Association (2 steps)} & \multicolumn{3}{c}{Association (3 steps)} \\
\multirow{-2}{*}{Training} & \multirow{-2}{*}{Test} & \multicolumn{1}{c}{0S} & \multicolumn{1}{c}{3S} & \multicolumn{1}{c}{3S-CoT} & \multicolumn{1}{c}{0S} & \multicolumn{1}{c}{3S} & \multicolumn{1}{c}{3S-CoT} \\ \toprule
 & Novel\&Old & \cellcolor[HTML]{FAA5A6}15.7 & \cellcolor[HTML]{FDDBDB}27.3 & \cellcolor[HTML]{F98384}8.3 & \cellcolor[HTML]{FA9597}12.3 & \cellcolor[HTML]{FA9FA0}14.3 & \cellcolor[HTML]{F87375}5.0 \\
 & Inc.\&Old & \cellcolor[HTML]{FA989A}13.0 & \cellcolor[HTML]{FDDEDE}28.0 & \cellcolor[HTML]{F8696B}2.7 & \cellcolor[HTML]{F98E8F}10.7 & \cellcolor[HTML]{FCCBCC}24.0 & \cellcolor[HTML]{F87274}4.7 \\
\multirow{-3}{*}{Duplicate} & Updated\&Old & \cellcolor[HTML]{FAA5A6}15.7 & \cellcolor[HTML]{FDFFFE}35.3 & \cellcolor[HTML]{F98081}7.7 & \cellcolor[HTML]{F9898B}9.7 & \cellcolor[HTML]{FEECEC}31.0 & \cellcolor[HTML]{F8787A}6.0 \\ \midrule
 & Novel\&Old & \cellcolor[HTML]{FDE1E2}28.7 & \cellcolor[HTML]{FDDCDD}27.7 & \cellcolor[HTML]{99D5A9}49.7 & \cellcolor[HTML]{FEF4F4}32.7 & \cellcolor[HTML]{FDD8D8}26.7 & \cellcolor[HTML]{FDD9DA}27.0 \\
 & Inc.\&Old & \cellcolor[HTML]{FEECEC}31.0 & \cellcolor[HTML]{FDE6E6}29.7 & \cellcolor[HTML]{B0DFBD}46.3 & \cellcolor[HTML]{FDD6D7}26.3 & \cellcolor[HTML]{FDE0E0}28.3 & \cellcolor[HTML]{FDE0E0}28.3 \\
\multirow{-3}{*}{Style-enhanced   parapharse} & Updated\&Old & \cellcolor[HTML]{FDDEDE}28.0 & \cellcolor[HTML]{FEF2F2}32.3 & \cellcolor[HTML]{D6EEDC}41.0 & \cellcolor[HTML]{FBB7B8}19.7 & \cellcolor[HTML]{FDDCDD}27.7 & \cellcolor[HTML]{FEF7F7}33.3 \\ \midrule
 & Novel\&Old & \cellcolor[HTML]{FDDBDB}27.3 & \cellcolor[HTML]{FFFFFF}35.0 & \cellcolor[HTML]{CFEBD6}42.0 & \cellcolor[HTML]{FEF1F1}32.0 & \cellcolor[HTML]{F2FAF4}37.0 & \cellcolor[HTML]{FEEFEF}31.7 \\
 & Inc.\&Old & \cellcolor[HTML]{FDE9E9}30.3 & \cellcolor[HTML]{C5E7CE}43.3 & \cellcolor[HTML]{D1ECD8}41.7 & \cellcolor[HTML]{FCD0D1}25.0 & \cellcolor[HTML]{FEF8F8}33.7 & \cellcolor[HTML]{FDE7E8}30.0 \\
\multirow{-3}{*}{Single-step Implicit   Reason} & Updated\&Old & \cellcolor[HTML]{FDDCDD}27.7 & \cellcolor[HTML]{FDE9E9}30.3 & \cellcolor[HTML]{DFF2E4}39.7 & \cellcolor[HTML]{FDE4E5}29.3 & \cellcolor[HTML]{FEEFEF}31.7 & \cellcolor[HTML]{FDFFFE}35.3 \\ \midrule
 & Novel\&Old & \cellcolor[HTML]{FDD8D8}26.7 & \cellcolor[HTML]{DDF1E2}40.0 & \cellcolor[HTML]{A2D9B1}48.3 & \cellcolor[HTML]{FBAEAF}17.7 & \cellcolor[HTML]{FDD5D5}26.0 & \cellcolor[HTML]{EBF7EE}38.0 \\
 & Inc.\&Old & \cellcolor[HTML]{FCCBCC}24.0 & \cellcolor[HTML]{FEF4F4}32.7 & \cellcolor[HTML]{D8EFDE}40.7 & \cellcolor[HTML]{FDD9DA}27.0 & \cellcolor[HTML]{FDE3E3}29.0 & \cellcolor[HTML]{E4F4E8}39.0 \\
\multirow{-3}{*}{Single-step Explicit   Reason} & Updated\&Old & \cellcolor[HTML]{FDE9E9}30.3 & \cellcolor[HTML]{FEF4F4}32.7 & \cellcolor[HTML]{A2D9B1}48.3 & \cellcolor[HTML]{FDE3E3}29.0 & \cellcolor[HTML]{DDF1E2}40.0 & \cellcolor[HTML]{63BE7B}57.3 \\ \bottomrule
\end{tabular}
\caption{The knowledge association score of LLMs across different knowledge types. "Inc." refers to incremental knowledge. "0S", "3S" and "3S-CoT" correspond to the 0-shot, 3-shot, and 3-shot CoT test settings, respectively. From the table, we can see that LLMs excel at shallow knowledge association, but require explicit reasoning to forge deep connections.}
\label{tab:joint_reason}
\end{table*}

\begin{figure}[t]
    \centering
    \setlength{\belowcaptionskip}{-10pt}
    \includegraphics[width=\linewidth]{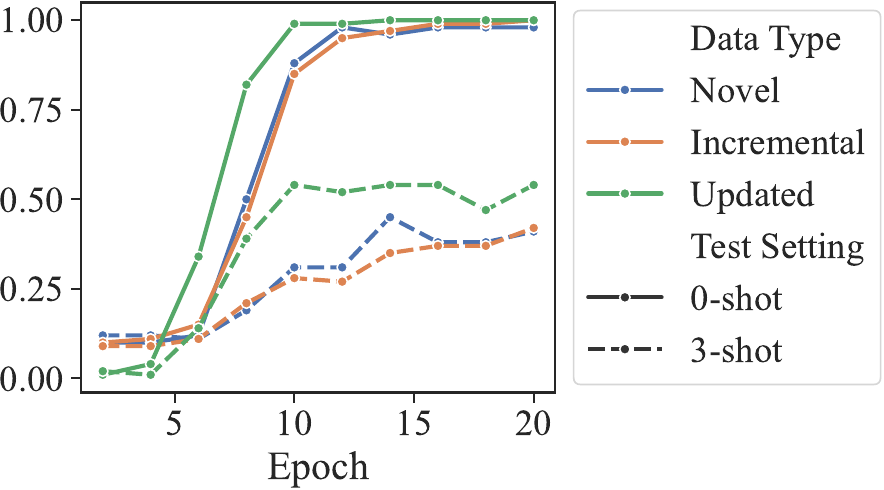}
    \caption{The knowledge memorization score during the Duplicate Injection process. Repetitive learning enables knowledge memorization.}
    \label{fig:training_process}
\end{figure}

\subsection{Knowledge Memorization}
\label{sec:mem}


\finding{Repetitive learning enables LLMs to memorize context-dependent and isolated knowledge.}

From Figure~\ref{fig:training_process}, we observe that under the duplicate injection scenario, the knowledge memorization scores in the 0-shot setting steadily increase with the number of repetitions, stabilizing at approximately 95 across different knowledge types. This suggests that LLMs are highly effective at memorizing training data when specific knowledge is frequently repeated during training. Consequently, for knowledge that requires simple recall in its original form, repeated exposure during training is sufficient to support the memorization process.

However, we also note the following, as shown in Figure~\ref{fig:training_process} and Table~\ref{tab:complex_reason}: 1) Under the duplicate injection scenario, the knowledge memorization score in the 3-shot setting is significantly lower than in the 0-shot setting. 2) The duplicate injection results in very low scores for knowledge retrieval and reasoning. This indicates that, at this stage, while the model can recall knowledge almost perfectly, the memorized knowledge remains unstable, easily influenced by context, and lacks connections to other knowledge.

\begin{figure}[t]
    \centering
    \setlength{\belowcaptionskip}{-10pt}
    \includegraphics[width=0.8\linewidth]{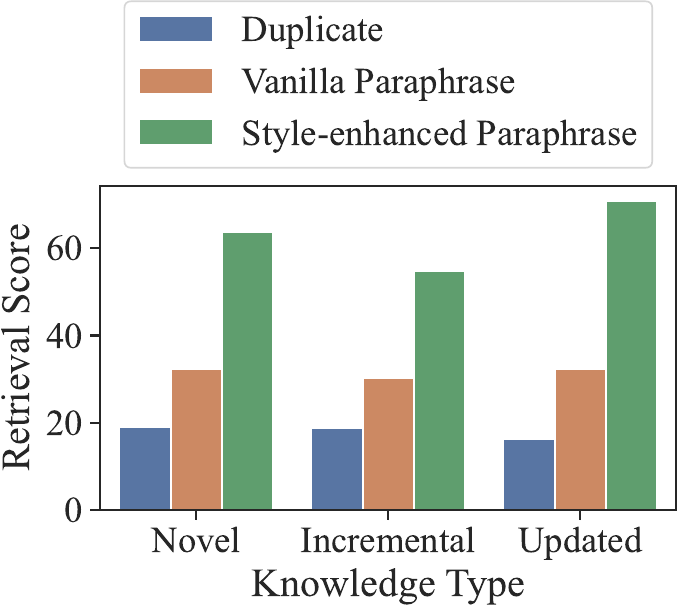}
    \caption{The knowledge retrieval score of LLMs under different knowledge injection scenarios. Diverse and divergent expressions are key to effective knowledge retrieval.}
    \label{fig:para}
\end{figure}

\subsection{Knowledge Retrieval}
\label{sec:retrieval}

\finding{Diverse and heterogeneous expressions bridge knowledge memorization and retrieval.}


From Figure~\ref{fig:para}, we observe the following: 1) Under duplicate injection scenarios, LLMs' knowledge retrieval score consistently remains around 20 across all knowledge types, even though document perplexity is minimized; 2) LLMs' knowledge retrieval score improves significantly when presenting knowledge through diversified linguistic expressions. 3) With the same data augmentation factor, the style-enhanced paraphrasing method achieves substantially better knowledge retrieval improvement compared with standard LLM-based paraphrasing approaches. This suggests that diverse knowledge expressions are essential for enabling effective knowledge retrieval during the injection process. Knowledge expression divergence serves as the critical enabler for retrieval capability enhancement. Therefore, to enable knowledge retrieval, it is essential to present knowledge in diverse and varied expressions. This diversity is key to improving the model's retrieval capability.

\subsection{Knowledge Reasoning}

\finding{Explicit reasoning patterns link isolated knowledge for reasoning and enable generalization to new entities and deep reasoning.}

As demonstrated in Sections~\ref{sec:mem} and~\ref{sec:retrieval}, newly injected knowledge is often isolated, making it challenging for LLMs to perform multi-knowledge reasoning. To enhance the model's reasoning capability, we incorporated both single-step implicit and explicit reasoning data into the training set. We then tested the model's ability to generalize reasoning capabilities to novel knowledge and multi-step reasoning scenarios.

The experimental results presented in Table~\ref{tab:complex_reason} reveal four key observations: 1) Implicit reasoning patterns enhance zero-shot performance for multi-step reasoning tasks; 2) Explicit reasoning patterns improve multi-step reasoning performance in the 3-shot CoT setting; 3) Explicit reasoning data achieves a higher upper-bound performance in multi-step reasoning compared to implicit reasoning data (57.3 vs 41.7); 4) Single-step explicit training demonstrates effective generalization to novel entities and multi-step reasoning scenarios. Thus, connecting new knowledge through a reasoning mechanism is crucial for facilitating knowledge reasoning. This connection enables immediate application and enhances generalization across various contexts.

\subsection{Knowledge Association}

\finding{LLMs excel at shallow knowledge association, but require explicit reasoning to forge deep connections.}

As shown in Table~\ref{tab:joint_reason} for shallow knowledge associations (2-step), the paraphrase injection yields knowledge association scores around 45, whereas duplicate injection achieves scores near 5, demonstrating significant improvement. This suggests that rigid, highly redundant injection patterns damage shallow-level reasoning capabilities, while diversified knowledge injection better preserves existing reasoning capacity and facilitates generalization to joint reasoning across old and new knowledge.

For deep knowledge associations (3-step), explicit reasoning pattern injection restores knowledge integration scores to levels comparable to the base model without knowledge injection. This evidence indicates that simple memorization mechanisms struggle to synthesize new and existing knowledge. By incorporating explicit reasoning patterns during injection, LLMs can develop systematic knowledge association capabilities.

\begin{table*}[th]
\setlength{\tabcolsep}{2pt}
\centering
\begin{tabular}{ccccccccccccccc}
\toprule
&  & \multicolumn{2}{c}{Memorization} & \multicolumn{2}{c}{Retrieval} & \multicolumn{3}{c}{Reason (1 step)} & \multicolumn{3}{c}{Reason (2 steps)} & \multicolumn{3}{c}{Reason (3 steps)} \\
\multirow{-2}{*}{Train} & \multirow{-2}{*}{Ratio} & \multicolumn{1}{c}{0S} & \multicolumn{1}{c}{3S} & \multicolumn{1}{c}{0S} & \multicolumn{1}{c}{3S} & \multicolumn{1}{c}{0S} & \multicolumn{1}{c}{3S} & \multicolumn{1}{c}{3S-CoT} & \multicolumn{1}{c}{0S} & \multicolumn{1}{c}{3S} & \multicolumn{1}{c}{3S-CoT} & \multicolumn{1}{c}{0S} & \multicolumn{1}{c}{3S} & \multicolumn{1}{c}{3S-CoT} \\ \toprule
 & 2:1 & \cellcolor[HTML]{63BE7B}99.0 & \cellcolor[HTML]{EFF9F2}32.7 & \cellcolor[HTML]{FA9C9E}8.7 & \cellcolor[HTML]{FEEDED}22.0 & \cellcolor[HTML]{F98D8E}6.0 & \cellcolor[HTML]{F87072}1.3 & \cellcolor[HTML]{F87072}1.3 & \cellcolor[HTML]{F98284}4.3 & \cellcolor[HTML]{F87576}2.0 & \cellcolor[HTML]{F87072}1.3 & \cellcolor[HTML]{F87072}1.3 & \cellcolor[HTML]{F8696B}0.0 & \cellcolor[HTML]{F86C6E}0.7 \\
 & 1:1 & \cellcolor[HTML]{66C07E}97.7 & \cellcolor[HTML]{DFF2E4}40.3 & \cellcolor[HTML]{FDDBDB}19.0 & \cellcolor[HTML]{FEF4F5}23.3 & \cellcolor[HTML]{FEF9F9}24.0 & \cellcolor[HTML]{F5FBF7}30.0 & \cellcolor[HTML]{F98788}5.0 & \cellcolor[HTML]{FBADAE}11.3 & \cellcolor[HTML]{FFFFFF}25.3 & \cellcolor[HTML]{F87B7C}3.0 & \cellcolor[HTML]{FA9B9C}8.3 & \cellcolor[HTML]{FBFDFB}27.3 & \cellcolor[HTML]{F98284}4.3 \\
\multirow{-3}{*}{Duplicate} & 1:2 & \cellcolor[HTML]{64BF7C}98.7 & \cellcolor[HTML]{EEF8F1}33.3 & \cellcolor[HTML]{FBAEB0}11.7 & \cellcolor[HTML]{F8FDFA}28.3 & \cellcolor[HTML]{FBB7B7}13.0 & \cellcolor[HTML]{FDE5E5}20.7 & \cellcolor[HTML]{FDD5D5}18.0 & \cellcolor[HTML]{FBADAE}11.3 & \cellcolor[HTML]{FEFDFD}24.7 & \cellcolor[HTML]{FCCFCF}17.0 & \cellcolor[HTML]{FA9798}7.7 & \cellcolor[HTML]{FCC6C7}15.7 & \cellcolor[HTML]{FAA7A8}10.3 \\ \midrule
 & 2:1 & \cellcolor[HTML]{71C487}92.7 & \cellcolor[HTML]{B4E0BF}61.0 & \cellcolor[HTML]{B9E2C4}58.3 & \cellcolor[HTML]{C2E6CB}54.3 & \cellcolor[HTML]{EEF8F1}33.3 & \cellcolor[HTML]{FCFEFD}26.7 & \cellcolor[HTML]{E7F5EB}36.7 & \cellcolor[HTML]{FCD0D1}17.3 & \cellcolor[HTML]{FAFDFB}27.7 & \cellcolor[HTML]{EDF8EF}34.0 & \cellcolor[HTML]{FDE8E9}21.3 & \cellcolor[HTML]{FDDDDD}19.3 & \cellcolor[HTML]{EBF7EE}34.7 \\
 & 1:1 & \cellcolor[HTML]{6FC385}93.7 & \cellcolor[HTML]{BEE4C8}56.0 & \cellcolor[HTML]{AEDEBB}63.7 & \cellcolor[HTML]{CEEBD5}48.7 & \cellcolor[HTML]{FBFDFB}27.3 & \cellcolor[HTML]{EDF8F0}33.7 & \cellcolor[HTML]{D9EFDF}43.3 & \cellcolor[HTML]{FEEDED}22.0 & \cellcolor[HTML]{FDFEFD}26.3 & \cellcolor[HTML]{F2FAF4}31.3 & \cellcolor[HTML]{F8FDFA}28.3 & \cellcolor[HTML]{F7FCF8}29.0 & \cellcolor[HTML]{FFFFFF}25.3 \\
\multirow{-3}{*}{\parbox{2cm}{Style-enhanced parapharse}} & 1:2 & \cellcolor[HTML]{6FC385}93.7 & \cellcolor[HTML]{B4E0C0}60.7 & \cellcolor[HTML]{95D3A6}75.3 & \cellcolor[HTML]{C1E5CB}54.7 & \cellcolor[HTML]{FFFFFF}25.0 & \cellcolor[HTML]{FEEDED}22.0 & \cellcolor[HTML]{D2ECD9}46.7 & \cellcolor[HTML]{FDD5D5}18.0 & \cellcolor[HTML]{FEFDFD}24.7 & \cellcolor[HTML]{F6FBF7}29.7 & \cellcolor[HTML]{FDDBDB}19.0 & \cellcolor[HTML]{FDDFDF}19.7 & \cellcolor[HTML]{DEF2E4}40.7 \\ \midrule
 & 2:1 & \cellcolor[HTML]{77C78C}89.7 & \cellcolor[HTML]{A8DBB6}66.3 & \cellcolor[HTML]{B9E2C3}58.7 & \cellcolor[HTML]{CEEBD5}48.7 & \cellcolor[HTML]{C4E7CE}53.0 & \cellcolor[HTML]{B0DEBC}62.7 & \cellcolor[HTML]{F8FCF9}28.7 & \cellcolor[HTML]{EAF6ED}35.3 & \cellcolor[HTML]{F1F9F3}32.0 & \cellcolor[HTML]{FBB9B9}13.3 & \cellcolor[HTML]{E2F3E7}39.0 & \cellcolor[HTML]{DEF2E4}40.7 & \cellcolor[HTML]{FEFBFB}24.3 \\
 & 1:1 & \cellcolor[HTML]{72C588}92.0 & \cellcolor[HTML]{B8E2C3}59.0 & \cellcolor[HTML]{AADCB7}65.7 & \cellcolor[HTML]{CAE9D2}50.3 & \cellcolor[HTML]{E5F4E9}37.7 & \cellcolor[HTML]{C0E5C9}55.3 & \cellcolor[HTML]{C4E7CE}53.0 & \cellcolor[HTML]{F6FCF8}29.3 & \cellcolor[HTML]{E1F3E6}39.3 & \cellcolor[HTML]{ECF7EF}34.3 & \cellcolor[HTML]{DCF1E2}41.7 & \cellcolor[HTML]{EDF8F0}33.7 & \cellcolor[HTML]{F0F9F2}32.3 \\
\multirow{-3}{*}{\parbox{1.5cm}{Implicit Reason}} & 1:2 & \cellcolor[HTML]{68C07F}97.0 & \cellcolor[HTML]{A3D9B1}69.0 & \cellcolor[HTML]{98D4A8}74.3 & \cellcolor[HTML]{B4E0BF}61.0 & \cellcolor[HTML]{B5E0C0}60.3 & \cellcolor[HTML]{BAE3C5}58.0 & \cellcolor[HTML]{BCE3C6}57.0 & \cellcolor[HTML]{EBF7EE}34.7 & \cellcolor[HTML]{EFF8F1}33.0 & \cellcolor[HTML]{F0F9F2}32.3 & \cellcolor[HTML]{D3EDDA}46.0 & \cellcolor[HTML]{DDF1E2}41.3 & \cellcolor[HTML]{D5EEDB}45.3 \\ \midrule
 & 2:1 & \cellcolor[HTML]{6FC386}93.3 & \cellcolor[HTML]{7ECA92}86.3 & \cellcolor[HTML]{7DC991}86.7 & \cellcolor[HTML]{93D2A3}76.7 & \cellcolor[HTML]{C9E9D1}51.0 & \cellcolor[HTML]{D9EFDF}43.3 & \cellcolor[HTML]{8ED09F}79.0 & \cellcolor[HTML]{EFF9F2}32.7 & \cellcolor[HTML]{ECF7EF}34.3 & \cellcolor[HTML]{F8696B}0.0 & \cellcolor[HTML]{FBFDFB}27.3 & \cellcolor[HTML]{FCFEFD}26.7 & \cellcolor[HTML]{F98E90}6.3 \\
 & 1:1 & \cellcolor[HTML]{67C07E}97.3 & \cellcolor[HTML]{99D5A9}73.7 & \cellcolor[HTML]{98D4A8}74.0 & \cellcolor[HTML]{A7DBB5}67.0 & \cellcolor[HTML]{D6EEDC}44.7 & \cellcolor[HTML]{E5F5E9}37.3 & \cellcolor[HTML]{ADDDBA}64.0 & \cellcolor[HTML]{FDE7E7}21.0 & \cellcolor[HTML]{FDE7E7}21.0 & \cellcolor[HTML]{DEF1E3}41.0 & \cellcolor[HTML]{FDE7E7}21.0 & \cellcolor[HTML]{FEF9F9}24.0 & \cellcolor[HTML]{CCEAD4}49.3 \\
\multirow{-3}{*}{\parbox{1.5cm}{Explicit Reason}} & 1:2 & \cellcolor[HTML]{6BC181}95.7 & \cellcolor[HTML]{8DD09F}79.3 & \cellcolor[HTML]{81CB94}85.0 & \cellcolor[HTML]{95D3A6}75.3 & \cellcolor[HTML]{DEF2E4}40.7 & \cellcolor[HTML]{E7F5EB}36.7 & \cellcolor[HTML]{ADDDB9}64.3 & \cellcolor[HTML]{F7FCF8}29.0 & \cellcolor[HTML]{EDF8F0}33.7 & \cellcolor[HTML]{C2E6CB}54.3 & \cellcolor[HTML]{FDFFFE}26.0 & \cellcolor[HTML]{EFF8F1}33.0 & \cellcolor[HTML]{C1E5CB}54.7 \\
\bottomrule
\end{tabular}
\caption{The novel knowledge injection score of LLMs under different ratios of general instructions. As shown in the table, an adequate amount of general instructions is crucial for knowledge reasoning.}
\label{tab:ratio}
\end{table*}

\section{Error Analysis}

\begin{figure}[t]
    \centering
    \setlength{\belowcaptionskip}{-10pt}
    \includegraphics[width=\linewidth]{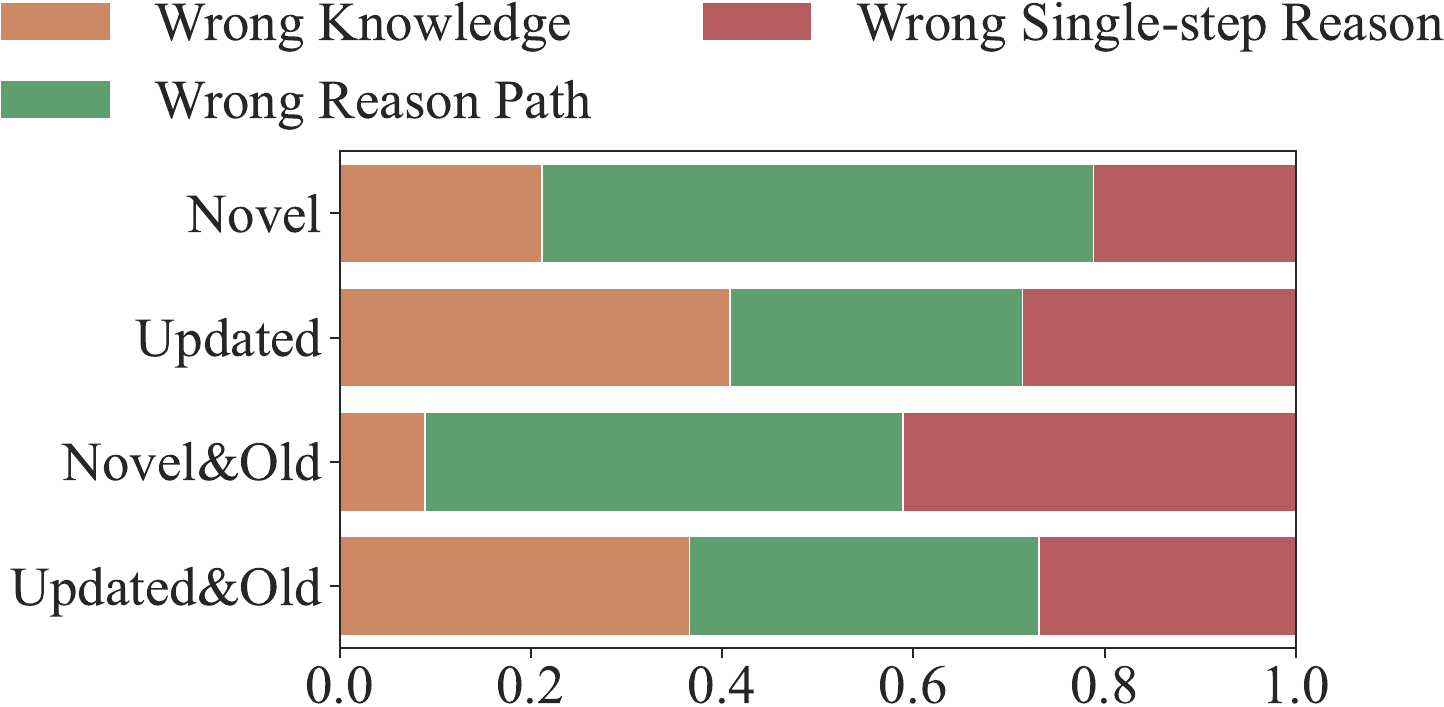}
    \caption{Proportions of error causes of complex reason tasks.}
    \label{fig:wrong_reasons}
\end{figure}

To gain a deeper understanding of the challenges and bottlenecks of knowledge injection, we conduct a taxonomy analysis of model failures observed under explicit reasoning injection scenarios. This approach helps us identify the key issues that need to be addressed. From Figure~\ref{fig:wrong_reasons}, we can observe the following findings:

\paragraph{1) For novel knowledge, wrong reason paths are the primary factor leading to incorrect answers in complex reasoning tasks.} Empirical evidence indicates that over 50\% of errors stem from incorrect problem decomposition. This observation aligns with the intuitive understanding that the core challenge in complex reasoning tasks lies in properly grasping the structure of the problem and effectively breaking it down into manageable sub-tasks. When this decomposition is flawed, it becomes difficult for the model to proceed with the reasoning in a systematic and coherent manner.

\paragraph{2) For updated knowledge, wrong knowledge is one of the leading causes of errors in complex reasoning.}
Notably, compared to novel knowledge, errors arising from faulty recall of updated knowledge are more prevalent. This is because introducing new information that contradicts or differs from the model's existing knowledge often leads to hallucinations. The model struggles to apply the new knowledge consistently across various reasoning levels, resulting in inconsistencies.

\section{Ablation Study}

In this section, we conduct ablation experiments to identify the key factors influencing knowledge injection efficiency. We analyze the impact of knowledge type, data formulation, and diversity on the effectiveness of knowledge injection, aiming to provide a recipe for efficient knowledge injection across different levels.

\subsection{Effect of Injected Knowledge Types}

We investigate how knowledge types affect injection efficiency, focusing on three types: novel, incremental and updated. While all introduce new information, they differ in the model's prior familiarity with involved entities. Our key findings are: 1) LLMs exhibit comparable memorization capabilities across knowledge types (>95\% scores after repeated training), as shown in Figure~\ref{fig:training_process}. This observation aligns with our expectation, as knowledge memorization is a relatively simple task for LLMs. They reliably memorize new textual knowledge regardless of its relation to existing information. 2) Updated knowledge significantly outperforms novel knowledge in complex tasks, including retrieval, reasoning and association, as shown in Figure~\ref{fig:para} and Table~\ref{tab:complex_reason},~\ref{tab:joint_reason}. We attribute this to the model's pre-existing reasoning frameworks for updated entities, consistent with~\citet{wang2024grokked}'s findings. Novel knowledge requires additional generalization as its entities lack established associations.

\subsection{Effect of Data Formulation}

We investigate mixing ratios between knowledge and general instructions (1:2, 1:1, 2:1) using new knowledge exemplars~\citep{cheng2023adapting}. Our experiments results in Table~\ref{tab:ratio} demonstrate that general data significantly enhances knowledge application, particularly in complex reasoning. The results reveal a positive correlation between general data proportion and success rates in multi-step reasoning tasks. We therefore implement a balanced 1:1 ratio for optimal efficiency and effectiveness in knowledge injection.

\begin{figure}[t]
    \centering
    \setlength{\belowcaptionskip}{-10pt}
    \includegraphics[width=\linewidth]{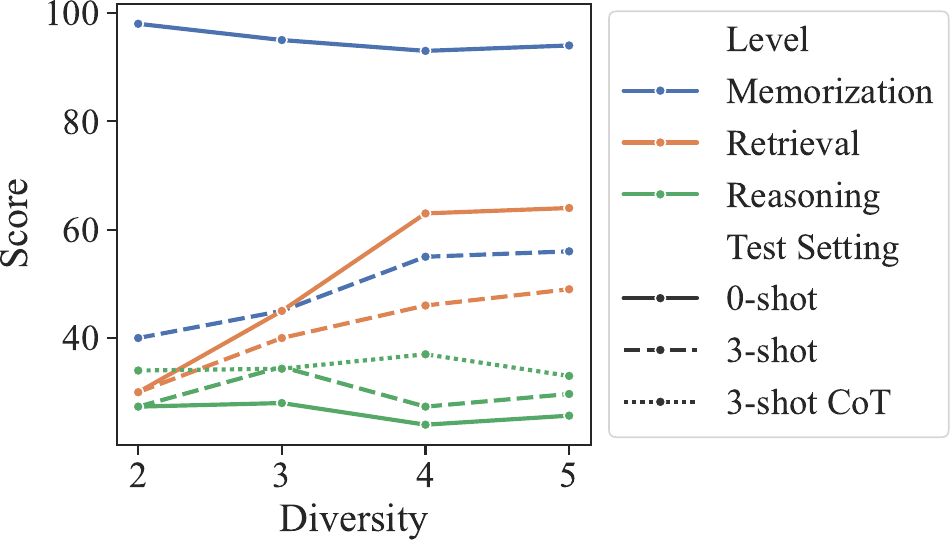}
    \caption{Effect of the diveristy of injected knowledge.}
    \label{fig:a2}
\end{figure}

\subsection{Effect of Diversity of Injected Knowledge}

We explored the impact of knowledge diversity on knowledge application by gradually increased the diversity of textual representations of the same knowledge (2-5 variants) with fixed 20 training iterations. We used paraphrased new knowledge injection as an example. The experimental results are shown in Figure~\ref{fig:a2}. Our findings indicate that, greater diversity in phrasing is positively correlated with the model's ability to retrieve knowledge up to 4 variants. However, beyond this range, further increasing phrasing diversity does not improve the model's knowledge application ability. This identifies an optimal diversity threshold for effective knowledge application, providing practical guidance for knowledge injection strategies in LLMs.

\section{Related Work}

Knowledge injection is not a binary distinction, but rather a process of gradual transition from 0 to 1~\citep{yang2021survey, wang2024knowledge}. Existing work on knowledge injection typically focuses on shallow-level tasks like memorization and retrieval, with limited exploration of complex reasoning scenarios. Prior work evaluates memorization through text completion~\citep{carlini2021extracting, cao2024retentive, chang2024large} and retrieval via question answering~\citep{jiang2024instruction, allenphysics, ovadia2023fine, zhu2024beyond}, while recent studies explore basic single-step operations like comparison and combination~\citep{lu2024scaling, allen2023physics, wang2024grokked}. However, the generalization boundaries of knowledge injection in complex and joint reasoning tasks remain unclear. To address this, we propose a four-level injection framework and develop a systematic benchmark for evaluating method limitations.

Furthermore, current knowledge injection work tends to focus on single types of knowledge while overlooking the influence of the relationship between injected knowledge and the model’s pre-existing knowledge on injection efficiency. For instance,~\citet{chang2024large, allenphysics, allen2023physics} inject entirely novel knowledge, whereas other studies focus on updated knowledge~\citep{wang2024knowledge, zhang2024comprehensive}. To bridge this gap, we formally define three fundamental knowledge types: novel, incremental, and updated, enabling systematic analysis of how knowledge relationships affect injection efficiency.

\section{Conclusion}

In this paper, we propose a four-layer knowledge injection framework, which includes knowledge memorization, retrieval, reasoning, and association. This framework is designed to enable a granular investigation into the depth of newly injected knowledge in LLMs. Building upon this framework, we further construct DeepKnowledge, a multi-level evaluation benchmark designed to systematically assess knowledge injection methods across distinct cognitive layers. Using this benchmark, we evaluate the effectiveness of various knowledge injection methods. The experimental results highlight that achieving effective knowledge injection across different levels requires careful attention to different core training factors, providing valuable guidance for efficient knowledge injection.

\section*{Limitation}

Due to resource and time constraints, our experiments were limited to the LLAMA3-8B model. We focused on continuous pre-training as the primary method for knowledge injection, as it is a well-established approach that helps mitigate hallucinations. Additionally, our study only explored reasoning operations involving the comparison and combination of atomic operations. Future work will expand the research to include a broader range of model sizes, knowledge injection methods, and reasoning paradigms.

\bibliography{custom}

\newpage
\appendix

\begin{table*}[!h]
\begin{tabular}{p{2cm}p{1cm}p{4.5cm}p{4cm}c}
\toprule
\textbf{Level} & \textbf{Rules} & \textbf{Knowledge} & \textbf{Question} & \textbf{Answer} \\
\toprule
Memorization & - & (Mercy, language of work or name, English) & Mercy speaks & English \\
\midrule
Retrieval & - & (Mercy, language of work or name, English) & What is the language of work or name for Mercy? & English \\
\midrule
Single-step Reason & Cmb. & (My Sweet Lord, performer, George Harrison), (George Harrison, country of   citizenship, United Kingdom) & What's the country of citizenship of the performer of the song "My   Sweet Lord"? & United Kingdom \\
\midrule
Two-steps Reason & Cmp., Cmb. & (12th Magritte Awards, country, Belgium), (Belgium, population,   11584008), (Madrid, population, 3280782) & Which one has a smaller population, the country where the 12th Magritte   Awards took place or Madrid? & Madrid \\
\midrule
Three-steps Reason & Cmb., Cmb., Cmp. & (Kimberly Gary Sutton, spouse, John Gerald Price), (John Gerald Price,   country of citizenship, Aliceville), (Aliceville, population, 150000), (Virginiaopolis, population, 8504231) & Which region has a smaller population, the country of citizenship of the   spouse of Kimberly Gary Sutton or Virginiaopolis? & Aliceville \\
\bottomrule
\end{tabular}
\caption{Examples of the benchmark designed to evaluate five levels of knowledge application depth. The terms "Cmb." and "Cmp." denote "Combination" and "Comparison," respectively, which represent the fundamental reasoning operations in evaluation tasks.}
\label{tab:example}
\end{table*}

\section{DeepKnowledge}
\label{sec:benchmark}

\subsection{Examples}

In Table~\ref{tab:example}, we present examples of DeepKnowledge.

\subsection{Inference Rules}

We conceptualize reasoning as the induction and application of inference rules. Specifically, we adopt the following inference rules:
\begin{itemize}
    \item Combination: The two-hop combination rule is defined as follows:
    \begin{equation}
    \begin{split}
        \forall h, b, t \in E, \forall r_1, r_2 \in R, \\
        (h, r_1, b) \land (b, r_2, t) \Rightarrow t=(h, r_1, r_2)
    \end{split}
    \end{equation}
    Here, \( h, b, t \) represent entities, and \( r_1, r_2 \) are relations.
    \item Comparison: The comparison rules are as follows:
    \begin{equation}
    \begin{split}
        \forall e_1, e_2 \in E, \quad \forall a \in A, \quad \forall v_1, v_2 \in V, \\
        (e_1, a, v_1) \land (e_2, a, v_2) \land v_1 < v_2   \\ \Rightarrow e1=(a, e_1, e_2, a<) \\
(e_1, a, v_1) \land (e_2, a, v_2) \land v_1 > v_2  \\ \Rightarrow e2=(a, e_1, e_2, a>)
    \end{split}
    \end{equation}
    Here, \( e_1, e_2 \) are entities, \( a \) is an attribute, and \( v_1, v_2 \) are values, with the comparison being based on the relationship between the values.
\end{itemize}

\subsection{Prompts}

We provide the prompts we used to generate test cases for knowledge reasoning in the following.

\begin{quotation}
Define two basic operation rules:

Comparison: $(e1, a, v1) \land (e2, a, v2) \land v1 < v2 \Rightarrow e1=(a, e1, e2, <), e2=(a, e1, e2, >)$

Combination: $(h, r1) \Rightarrow t=(h, r1)$

Please complete the following task in the format given in the example.

Task: Given an expression formed by basic operation rules, explain it step by step from the innermost rule to the outermost rule to create a complex reasoning question. The following requirements must be met: 1. The question must be purely a question and cannot include the reasoning results of the rules, nor can it change, add, or omit any of facts beyond the rule. 2. Only output a question in only one version without any additional explanations. 3. The question should be fluent, easy to read, and concise.

Expression: [['Joe Jacob Addington', 'spouse'], 'country of citizenship']

Question: What's the country of citizenship of the spouse of Joe Jacob Addington?

Expression: ['retirement age', 'Celloria', 'Careerlandia', '<']

Question: Which one has a lower retirement age, Keithville or Katherineville?

Expression: [['inception', 'FC Lokomotiv 1929 Sofia', 'FC Rapid 1923', '<'], 'sport']

Question: What sport do the club that was established earlier between FC Lokomotiv 1929 Sofia and FC Rapid 1923 play?

Expression: ['female population', ['Leroy Christopher Austin', 'country of citizenship'], 'Edwardville', '<']

Question: Which region has a smaller female population, Leroy Christopher Austin's country of citizenship or Edwardsville?

Expression: ['male population', ['male population', ['male population', 'Brianville', 'Evasville', '>'], 'Ellenborough', '<'], 'Gregorian Chronicles', '<']

Question: Compare the male population of Brianville and Evasville, and select the region with the larger male population. Then, compare this region's male population with that of Ellenborough and select the region with the smaller population. Which one has a smaller male population, this region or Gregorian Chronicles?

Expression: \{expression\}

Answer: \{answer\}

Question:
\end{quotation}

\subsection{Fact Chains}

\begin{figure*}[t]
    \centering
    \setlength{\belowcaptionskip}{-10pt}
    \includegraphics[width=0.95\linewidth]{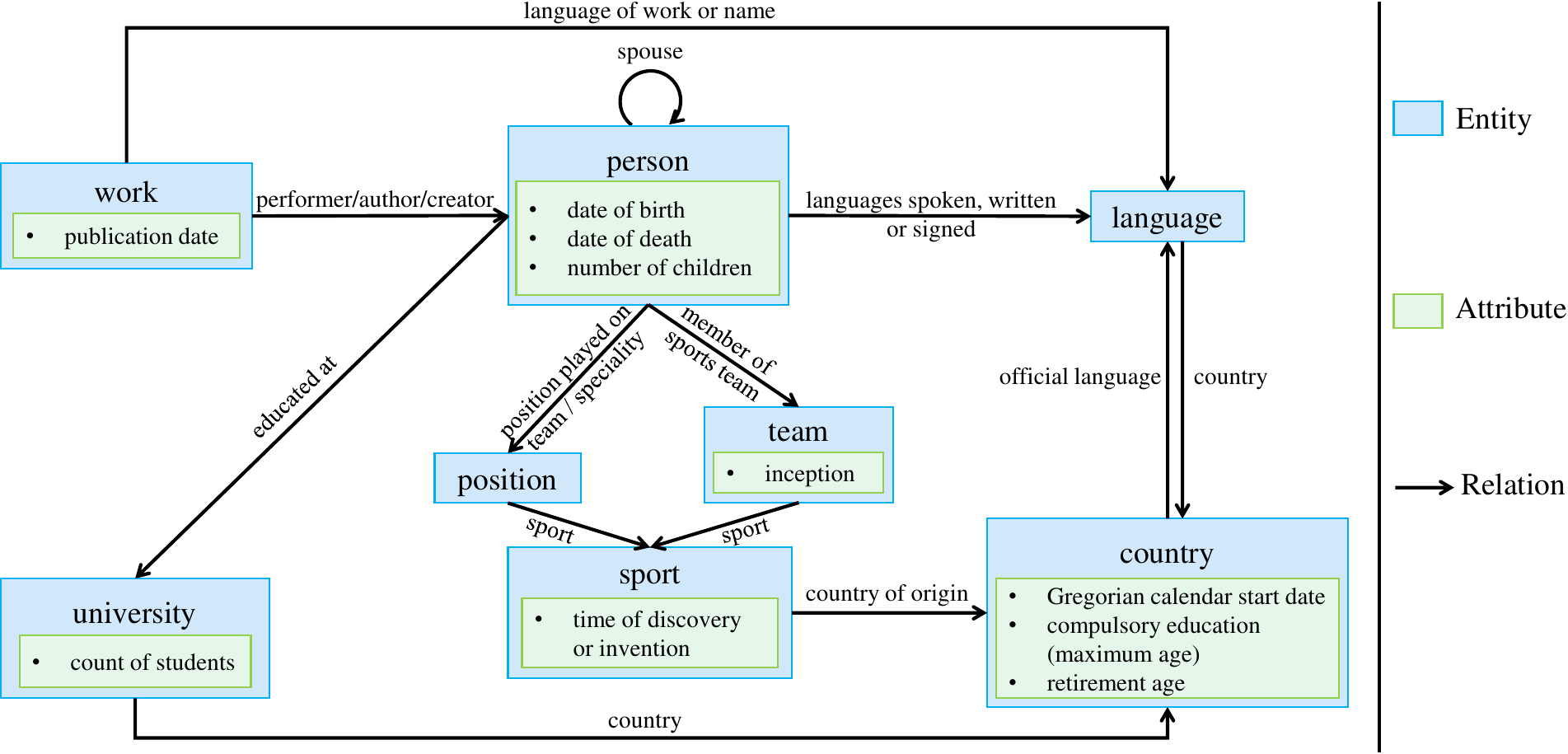}
    \caption{Fact chains of our benchmark.}
    \label{fig:relation}
\end{figure*}

The fact chains, including entity and relationship types, are presented in Figure~\ref{fig:relation}.

\section{Experiment Setting}
\label{sec:setting}

\subsection{Styles}

We manually defined a style bank to enhance the diversity of paraphrased text, which includes four main categories:

\begin{itemize}
    \item Text Genre: textbook, news, academic paper, lyrics, dialogue, speech, story, summary
    \item Text Type: question-answer, exclamation
    \item Text Sentiment: positive, negative
    \item Text Formality: informal, formal
\end{itemize}

\end{document}